\documentclass[runningheads]{llncs}
\usepackage[T1]{fontenc}
\usepackage{graphicx}
\begin{document}
\newcommand{\gbf}[1] {\mbox{\boldmath${#1}$\unboldmath}}
\newcommand{\beq}{\begin{equation}}
\newcommand{\eeq}{\end{equation}}
\newcommand{\bea}{\begin{eqnarray}}
\newcommand{\eea}{\end{eqnarray}}
\newcommand{\R}{\hbox{I \kern -.5em R}}

\title{The Twist Decomposition of Serial Robots\\ Under Lower-Mobility Tasks}
\author{Luc Baron\inst{1}  \and Damien Chablat\inst{2}}
\authorrunning{L. Baron and D. Chablat}
\institute{Luc Baron\inst{1}  \and Damien Chablat\inst{2}}
\institute{Polytechnique Montréal, box 6079, sta. CV, QC, H3C 3A7 Canada
\email{Luc.Baron@polymtl.ca} \and
Université de Nantes, Ecole Centrale de Nantes, LS2N, UMR CNRS 6004, \\ 1 rue de la Noë, 44321 Nantes, France,
\email{Damien.Chablat@cnrs.fr}}
%
\maketitle

\abstract{This paper introduces a twist decomposition framework for serial manipulators performing lower mobility tasks. Rather than relying on Jacobian null-space projections, the method separates the end-effector twist into task and redundant components using geometrically defined twist projectors. This formulation provides a direct and intuitive distinction between task-relevant and task-irrelevant motions in operational space, enabling a compact inverse kinematics scheme that naturally handles both manipulator and task redundancy.
}

\section{Introduction}
\label{sec:1}
Serial robotic manipulators are widely used in industry for material handling as well as for manufacturing processes such as polishing, welding, deburring, and many others. When such tasks of lower-mobility are executed using a full six-degree-of-freedom robot, the system becomes functionally redundant \cite{Baron2007}. When the number of joints exceeds the dimension of the end-effector (EE) operational space, the manipulator is redundant for all tasks, i.e., intrinsically redundant.
The mapping between joint-space motion and EE motion is classically described by the Jacobian of the manipulator, which relates joint velocities to EE twists in operational space \cite{Liegeois1977}. Intrinsic redundancy is usually resolved using the generalized inverse of a rectangular Jacobian matrix combined with gradient-projection methods \cite{Nakamura1991,Colome2012}. However, this approach requires a rectangular Jacobian with more columns than rows in order to exploit a null space. As a result, these methods cannot be directly applied to functional redundancy \cite{Angeles2015}.
The constraints imposed by lower-mobility tasks reduce the effective mobility of the EE. In such situations, only a subset of the six components of the EE twist are relevant to the task, leading to redundancy of the task. The distinction between manipulator redundancy and task redundancy, as well as their unified treatment in operational space, has been investigated in several works. More recent studies have extended these concepts to hierarchical task formulations and constraint-based control, where multiple task objectives are combined through projection operators or weighted pseudo-inverses \cite{Khatib2006,Zlajpah2017,Flacco2015}.

However, most existing methods rely on explicit null-space projectors constructed from the Jacobian. These approaches become cumbersome when task constraints vary along the trajectory or when the translational and rotational components of the task are selectively constrained. Instead, geometric approaches have been proposed to directly decompose the twist of EE \cite{Angeles2015,Huo2011}. This work presents a unified formulation that resolves both manipulator and task redundancy within a task-priority hierarchy. The proposed framework accommodates both constant and time-varying task constraints.

The remainder of this paper is organized as follows. Section 2 reviews the kinematic model and defines the operational and task spaces. Section 3 introduces the proposed twist decomposition framework. Section 4 demonstrates its effectiveness through numerical examples that involve a planar 3R manipulator and an industrial deburring task. Finally, Section 5 concludes the paper with a broader discussion.

\section{Kinematic Model}
\label{sec:2}
For a serial manipulator, the {\em Jacobian} matrix ${\bf J}$ maps the joint velocities $\dot{\gbf \theta}$
to the twist of the end-effector (EE) $ {\bf t}$, i.e., from the joint-velocity space ${\cal J}$ to
the operational space ${\cal O}$, as
\begin{equation}
 {\bf t} = {\bf J} \dot{\gbf \theta},\quad
 {\bf t} \equiv \left[ \begin{array}{c} {\gbf \omega} \\ {\bf v}\end{array} \right] \in SE(3),\quad
 \dot{\gbf \theta} \equiv \left[ \begin{array}{c} \dot{\theta}_1 \\ \vdots\\ \dot{\theta}_n\end{array} \right] \in  {\cal J} .
\label{t}
\end{equation}
Here, ${\gbf \omega}$ denotes the angular velocity of the EE and
${\bf v}$, the linear velocity of a reference point on the EE.
The operational space ${\cal O}$, the task space ${\cal T}$, and
the joint space ${\cal J}$ satisfy
\begin{equation}
6 \ge {\rm dim}({\cal O}) = n_o
   \ge {\rm dim}({\cal T})  = n_t
   \le {\rm dim}({\cal J}) = n .
\end{equation}

\noindent{\bf Example 2.1}: For a planar $2$R manipulator, ${\bf J} \in \R^{3 \times 2}$.
The EE twist has 3 components; hence $n_o = 3$ with $n=2$ joints.
The manipulator is overconstrained and cannot generate an arbitrary planar twist.

The inverse kinematic (IK) problem is typically solved iteratively by mapping the twist error ${\gbf \Delta}{\gbf t}$
to a joint update ${\gbf \Delta}{\gbf \theta}_k$, which is likely to reduce the task error in step $k+1$ with
\begin{equation}
{\gbf \theta}_{k+1}= {\gbf \theta}_k+{\gbf \Delta}{\gbf \theta}_k .
\end{equation}

\noindent{\bf No Redundancy}: When $n_o = n$, ${\bf J}$ is square. The IK is computed with
\begin{equation}
{\gbf \Delta}{\gbf \theta}_k = {\bf J}^{-1} {\gbf \Delta}{\gbf t}, \label{inverse}
\end{equation}
provided that ${\bf J}$ is full rank.
\vskip 0.1in

\noindent{\bf Manipulator Redundancy}:  When $n_o < n$, ${\bf J}$ is rectangular.
The IK is computed with
\begin{equation}
{\gbf \Delta}{\gbf \theta}_k
   = \underbrace{({\bf J}^\dag) {\gbf \Delta}{\gbf t}}_{\rm primary~task}
   + \underbrace{({\bf 1} - {\bf J}^\dag {\bf J}){\bf h}}_{\rm secondary~objective}, \quad
  {\mathbf J}^\dag \equiv {\mathbf J}^T({\bf J} {\bf J}^T)^{-1},  
\label{general}
\end{equation}
where ${\bf h}$ is an arbitrary vector in the joint space.
The first term of (\ref{general}) yields the minimum-norm solution of (\ref{t}) and
accomplishes the primary task,
while the second term produces self-motion in the null space of ${\bf J}$
allowing secondary objectives to be satisfied without affecting the primary task.
The redundancy of the manipulator is defined as $r_m = n - n_o$.

\noindent{\bf Example 2.2}: For a planar $4$R manipulator, ${\bf J} \in \R^{3 \times 4}$,
with $n_o=3$ and $n=4$ joints.
The manipulator is always redundant $r_m = 1$, while it can generate an arbitrary planar twist.

\noindent{\bf Redundancy of tasks}: When $n_t < n_o \le n$, only a subset of the EE twist components is required to perform the task.
The Jacobian is reduced to ${\bf J} \in \R^{n_t \times n}$. The redundancy of tasks is defined as $r_t = n_o - n_t$.

\noindent{\bf Example 2.3}: For a planar $3$R manipulator, ${\bf J} \in \R^{3 \times 3}$.
If the task is end-point positioning without orientation control,  the rotational component $\omega_z$ is removed, resulting in a $2 \times 3$ Jacobian and a redundancy of the task $r_t = 1$.

\subsection{Decomposition of vectors}
\label{sec:3.1}
Any vector $(\,\cdot\,) \in \R^3$, such as ${\gbf \omega}$ or ${\bf v}$,
can be decomposed into two orthogonal components: one lying in the
subspace ${\cal M}$, denoted $[\,\cdot\,]_M$, and one
lying in the orthogonal subspace ${\cal M}^\perp$, denoted $[\,\cdot\,]_{M^\perp}$,
using the associated projectors ${\bf M}$ and ${\bf M}^\perp$,
\begin{eqnarray}
(\,\cdot\,) &=& [\,\cdot\,]_M + [\,\cdot\,]_{M^\perp} 
            = {\bf M} (\,\cdot\,) + {\bf M}^\perp (\,\cdot\,)
            = ({\bf M} + {\bf M}^\perp) (\,\cdot\,) \label{decomposition}
\end{eqnarray}
The projectors satisfy ${\bf M} + {\bf M}^\perp = {\bf 1}$ and ${\bf M} {\bf M}^\perp = {\bf O}$,
where ${\bf 1}$ and ${\bf O}$ denote the identity and zero matrices $3 \times 3$.
For subspaces of $\R^3$, the projectors take the forms
\begin{equation}
{\bf M} = \left\{ \begin{array}{c}
{\bf 1} ~~ ({\rm rank}=3) \\
{\bf P} ~~ ({\rm rank}=2) \\
{\bf L} ~~ ({\rm rank}=1) \\
{\bf O} ~~ ({\rm rank}=0)
\end{array} \right. ,\quad
{\bf M}^\perp = \left\{ \begin{array}{c} {\bf O} \\ {\bf L} \\
{\bf P} \\ {\bf 1} \end{array} \right. ,
\label{Mi}
\end{equation}
where ${\bf P} \equiv {\bf 1} - {\bf e} {\bf e}^T$ is the projector onto a plane of normal unit vector ${\bf e}$,
and ${\bf L} \equiv {\bf e} {\bf e}^T$ is the projector onto a line spanned by ${\bf e}$.
These matrices satisfy the standard projector properties:
{\em symmetry}, {\em idempotency}, and {\em singularity} (except for {\bf 1}).

\subsection{Decomposition of twists}
\label{sec:3.2}
Any twist $(\,\cdot\,) \in SE(3)$ can be decomposed into components that lie in
the operational subspace (or task) ${\cal O}$ and its orthogonal complement ${\cal O}^\perp$ 
using twist projectors ${\bf T}$ and ${\bf T}^\perp$ :
\begin{eqnarray}
(\,\cdot\,) &=& [\,\cdot\,]_{\cal O} + [\,\cdot\,]_{{\cal O}^\perp}
            = {\bf T} (\,\cdot\,) + {\bf T}^\perp (\,\cdot\,) 
            = ({\bf T} + {\bf T}^\perp) (\,\cdot\,) \label{decomposition_t}
\end{eqnarray}
These projectors are block-diagonal matrices built from projectors of (\ref{Mi}):
\begin{eqnarray}
{\bf T} &\equiv& \left[ \begin{array}{cc}
{\bf M}_\omega & {\bf O} \\
{\bf O}  & {\bf M}_v \end{array} \right],\quad
{\bf T}^\perp \equiv {\bf 1} - {\bf T} = \left[ \begin{array}{cc}
{\bf 1} - {\bf M}_\omega & {\bf O} \\
{\bf O} & {\bf 1} - {\bf M}_v \end{array} \right] .
\end{eqnarray}
Thus, any twist decomposes as
\begin{equation}
 {\bf t} = {\bf t}_{\cal O} + {\bf t}_{{\cal O}^\perp}
         = {\bf T t} + ({\bf 1} - {\bf T}) {\bf t}
 \label{twist_decomp}
\end{equation}

\section{Twists Decomposition Approach}
For any tasks, the instantaneous EE twist can be decomposed into the task subspace ${\cal T}$
and a component lying in the redundant subspace ${\cal T}^\perp$. Substituting (\ref{twist_decomp}) into
(\ref{inverse}) yields
\begin{equation}
{\mathbf \Delta}{\mathbf \theta}^k
 = \underbrace{({\bf J}^\dag {\bf T}) {\mathbf \Delta t}}_{\rm primary~task}
 + \underbrace{{\bf J}^\dag ({\bf 1}-{\bf T}){\bf J}{\bf h}}_{\rm secondary~task} ,
\label{liguo}
\end{equation}
where ${\bf h}$ is an arbitrary joint-space vector.
The first term produces joint displacements required to perform the primary task,
while the second term generates joint motions in the redundant subspace that do not affect the primary task.
Unlike the classical approach of (\ref{general}), this formulation does not require explicit projection
onto the null-space of ${\bf J}$; instead, it directly decomposes the EE twist according to the task.

For illustration purposes, let us avoid joint limits by keeping ${\gbf \theta}$ close to the mid-joint position $\bar{\gbf \theta}$ such as
\begin{equation}
z = \frac{1}{2} (\bar{\gbf \theta} - {\gbf \theta})^T {\bf W}^T {\bf W}  (\bar{\gbf \theta} - {\gbf \theta}) \rightarrow  \min
\label{z}
\end{equation}
with $\bar{\gbf \theta}$ and ${\bf W}$ being defined as
\begin{equation}
\bar{\gbf \theta} \equiv \frac{1}{2} ({\gbf \theta}_{max} + {\gbf \theta}_{min}), \quad
\underline{\gbf \theta} \equiv \frac{1}{({\gbf \theta}_{max} - {\gbf \theta}_{min})}, \quad
{\bf W} \equiv  {\rm\bf diag}(\underline{\gbf \theta})
\end{equation}
The gradient ${\gbf \nabla}z$ is a vector in the joint-space that points toward increasing $z$.
Therefore, to minimize $z$, we chose ${\bf h} = -{\gbf \nabla}z = {\bf W} (\bar{\gbf \theta} - {\gbf \theta})$.

\section{Numerical Examples}
\label{sec:4}
The following examples highlight redundancy under lower-mobility tasks and demonstrate the applicability of the proposed framework to constant and time-varying redundant tasks.
In both examples, the redundancy can be computed from the rank of twist projectors as
\begin{equation}
r_t = {\rm rank}({\bf T}_{\cal O}) -  {\rm rank}({\bf T}_{\cal T}) = 1
\end{equation}

\begin{figure}[hbt]
\centering
\includegraphics[scale=0.18]{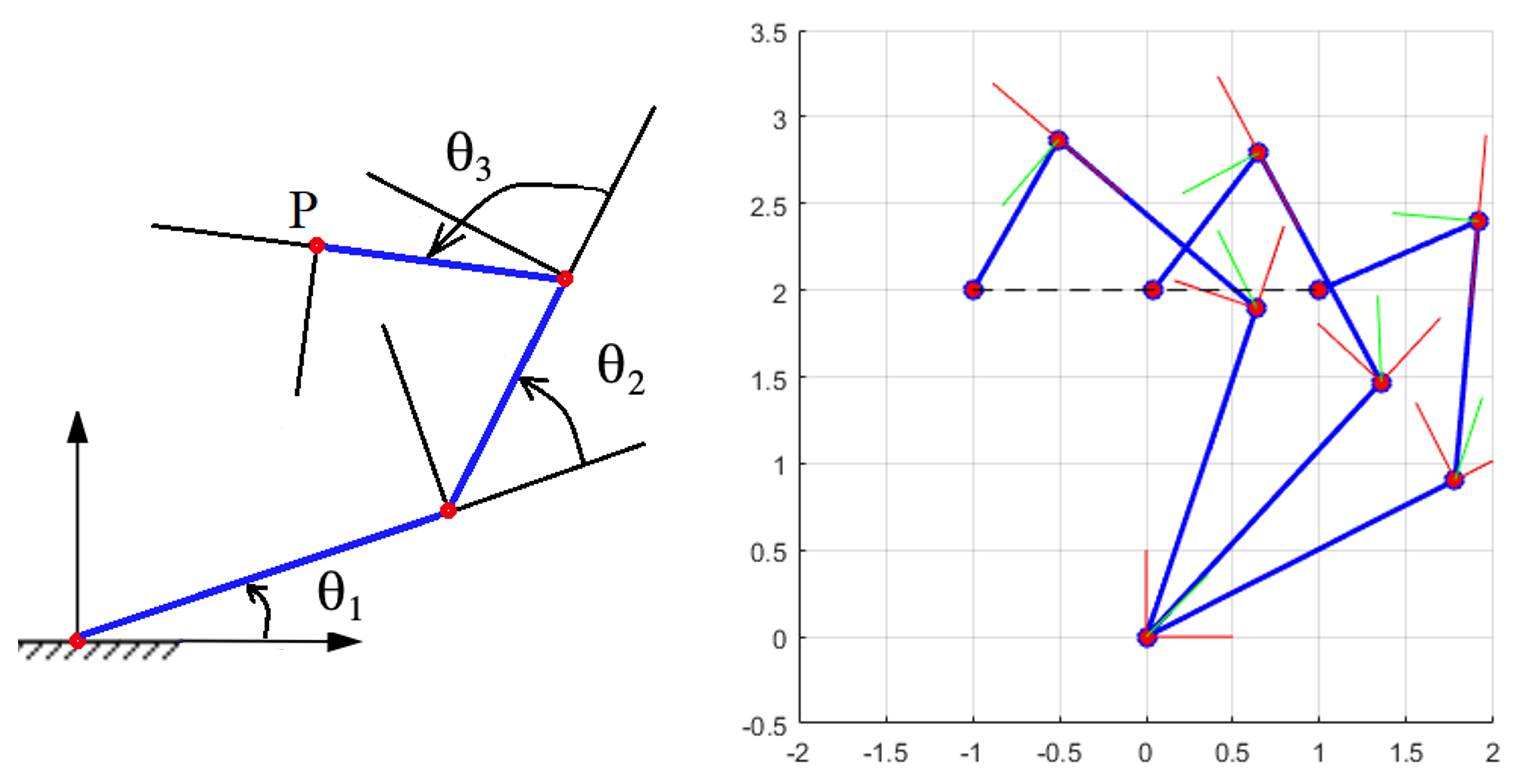}
  \begin{picture}(0,0)
 	\put(-110,95){$1$}
	\put( -35, 95){$2$}
	\put(-235, 42){$L_1$}
	\put(-195, 78){$L_2$}
	\put(-214,100){$L_3$}
	\put(-210, 22){$x_0$}
	\put(-275, 70){$y_0$}                 
	\put(-158, 60){$x_1$}
	\put(-217, 78){$y_1$} 
	\put(-165,125){$x_2$}
	\put(-213,114){$y_2$} 
 	\put(-255, 94){$x_3$}
	\put(-234, 70){$y_3$} 
        \put(-294,  -7){(a) Geometry of the 3R manipulator}
        \put(-128,  -7){(b) Straight line task from 1 to 2} 
        \put( -290,  10){${\gbf \theta}_{min} = [ 0 ~ 0 ~ 0]^T$}
        \put( -225,  10){${\gbf \theta}_{max} = [ \pi/2 ~ \pi/2 ~ \pi]^T$} 
        \put( -180,  45){$L_1 = 2$} 
        \put( -180,  35){$L_2 = 1.5$} 
        \put( -180,  25){$L_3 = 1$}         
  \end{picture}
\caption{Planar 3R manipulator performing a straight line}
\label{fig1}   
\end{figure}
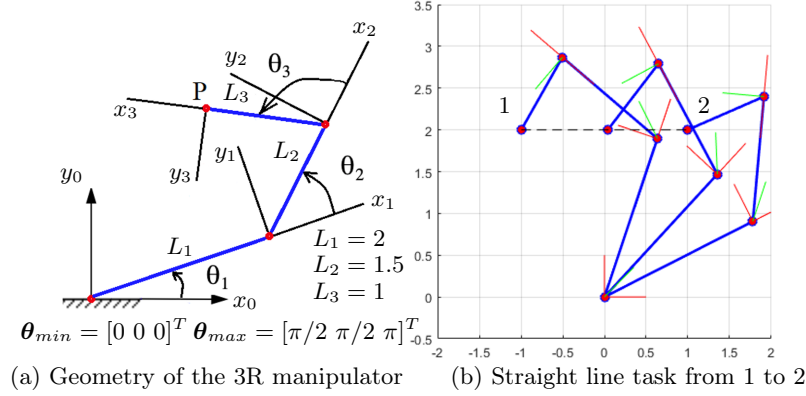

\subsection{Planar $3$R manipulator under positioning tasks}
\label{sec:4.1}
 Consider a planar 3R manipulator performing an end-point positioning task, for which the redundant motion subspace
remains constant throughout the workspace with ${\bf e} = [0 ~ 0 ~ 1]^T$. The twist decomposes for ${\cal O}$ and ${\cal T}$ as
\begin{eqnarray}
{\bf T}_{\cal O} &=& \left[ \begin{array}{cc}
   {\bf e} {\bf e}^T & {\bf O} \\
   {\bf O}  & ({\bf 1} - {\bf e} {\bf e}^T) \end{array} \right],\quad
{\bf T}_{\cal T} = \left[ \begin{array}{cc}
   {\bf O}  & {\bf O} \\
   {\bf O} & ({\bf 1} - {\bf e} {\bf e}^T) \end{array} \right] .
\end{eqnarray}
Figure \ref{fig1} illustrates the geometry of the manipulator and the trajectory of the end-point.
Figure \ref{fig4} presents the joint-space roadmap for a motion from point $P_1$ to point $P_2$.
The green curve corresponds to the minimum-norm solution, which satisfies the primary task only, while the blue curve shows the solution obtained when joint-limit avoidance is incorporated as a secondary objective. Although not required for the computation, the black vertical lines represent the self-motion manifold associated with the task, as discussed in \cite{Albu2023}.

\subsection{Fanuc M16iB manipulator under deburring tasks}
\label{sec:4.2}
\begin{figure}[b]
\centering
\includegraphics[scale=0.50]{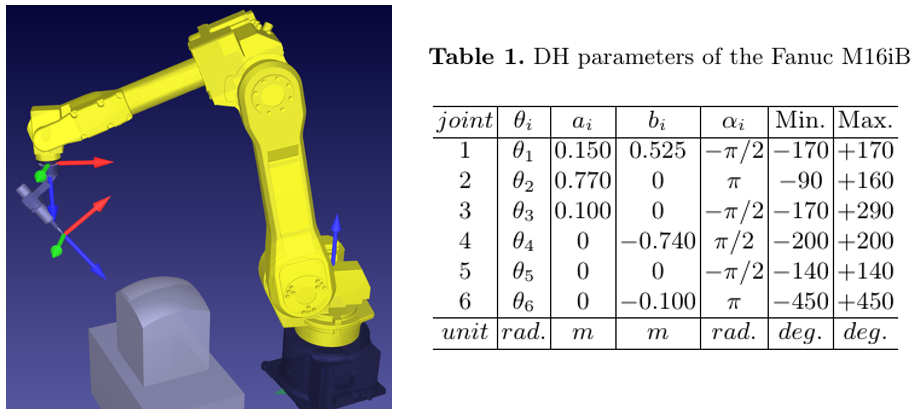}      
\caption{Fanuc M16iB robot with tool (axis x=red, y=green, z=blue)}
\label{fig2}   
\end{figure}

Consider an industrial robot with six-axes as Fanuc M16iB performing a deburring task,
for which the redundant motion is a rotation around the tool axis, shown in blue in Fig.~\ref{fig2}.
The geometry of the tool is
\begin{equation}
 {\bf R}_{tool} =  \left[ \begin{array}{cccc}
   \cos\phi & 0 & \sin\phi \\
   0  & 1 & 0 \\
   -\sin\phi & 0 & \cos\phi \end{array} \right],\quad
 {\bf p}_{tool} = \left[ \begin{array}{c} 0.032 \\ 0 \\ 0.260 \end{array} \right]~m,\quad
 \phi = 48^\circ.
\end{equation}
The robot is used to deburre the four edges of a spherical part
of $0.25 \times 0.25$ m of radius $0.3$ m and centered at the coordinate $(0, ~1, ~0.3)~m$.
The twist is decomposed as
\begin{eqnarray}
{\bf T}_{\cal O} &=& \left[ \begin{array}{cc}
   {\bf 1} & {\bf O} \\
   {\bf O}  & {\bf 1} \end{array} \right],\quad
{\bf T}_{\cal T} = \left[ \begin{array}{cc}
   ({\bf 1} - {\bf e} {\bf e}^T)  & {\bf O} \\
   {\bf O} & {\bf 1} \end{array} \right], \quad
{\bf e} = {\bf R}_{robot} {\bf R}_{tool} {\bf k},
\end{eqnarray}
where ${\bf R}_{robot}$ is the orientation of the flange frame relative to the robot base frame
and ${\bf k} \equiv [0 ~ 0 ~ 1]^T$. Figure \ref{fig2} shows the DH parameters of the manipulator and the geometry of the tool.

\begin{figure}[bht]
\centering
\includegraphics[scale=0.55]{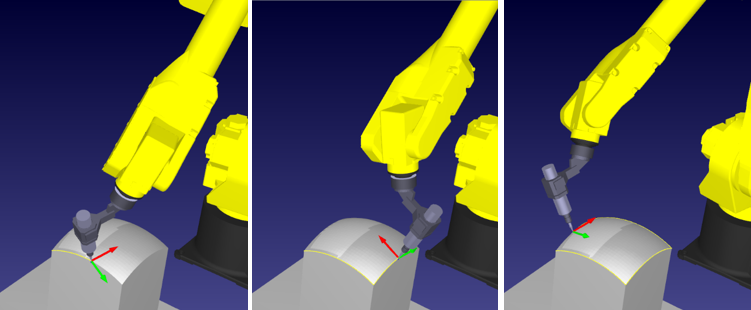}
\includegraphics[scale=0.40]{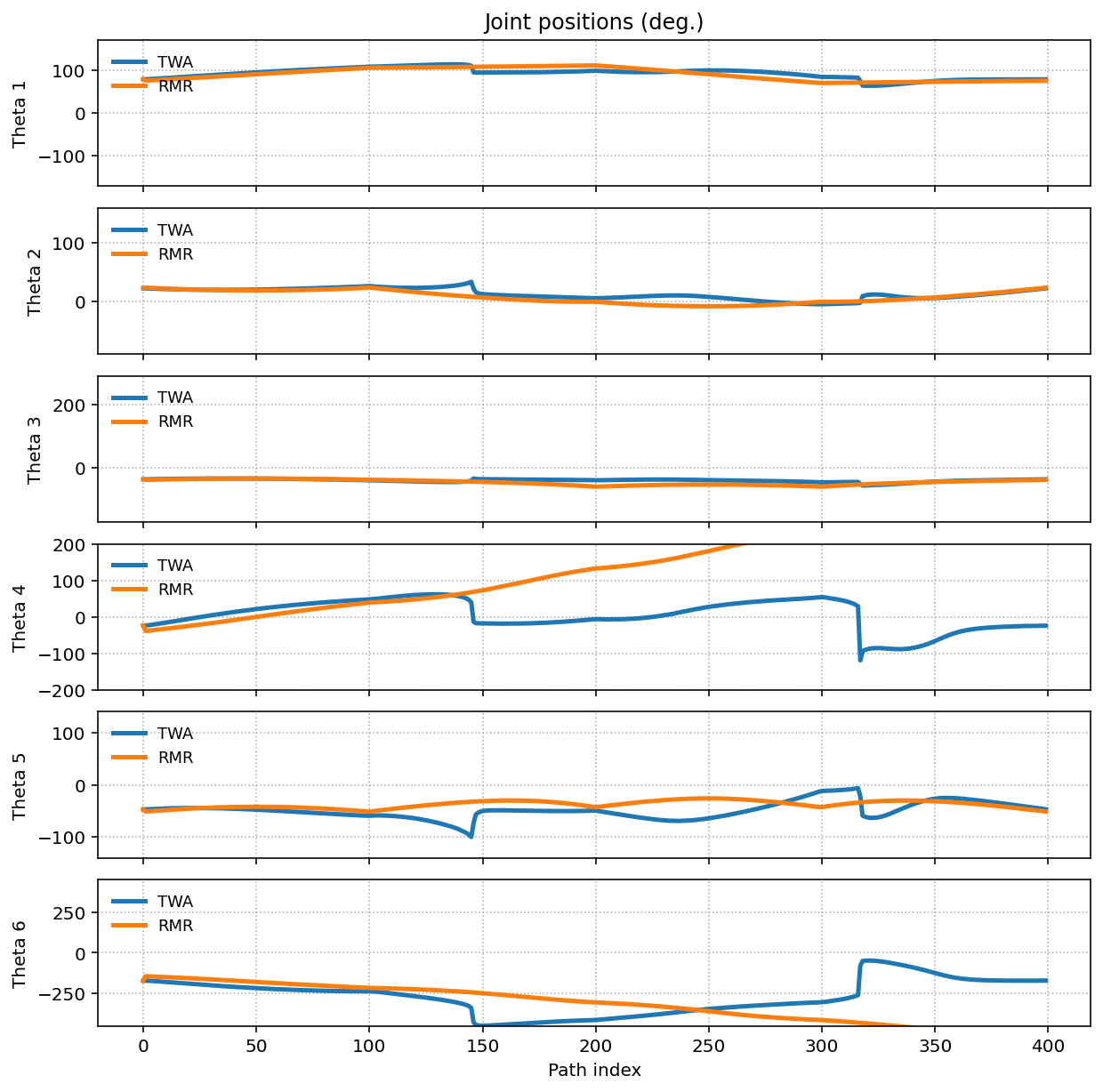}
\caption{Joint positions along the four edges (100 points per edge) with the twist decomposition framework (TWA: blue) and with the resolved-motion rate (RMR: orange)}  
\label{fig3}   
\end{figure}

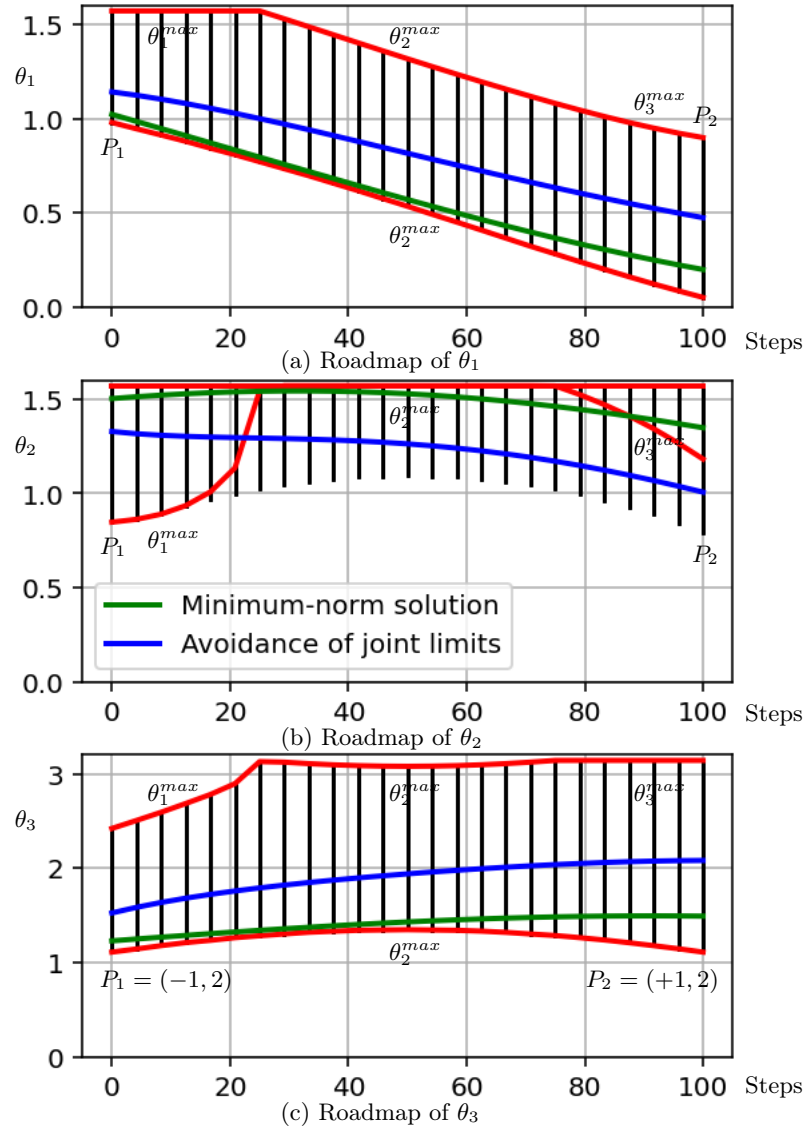
\begin{figure}[hbt]
\centering
\includegraphics[scale=1.0]{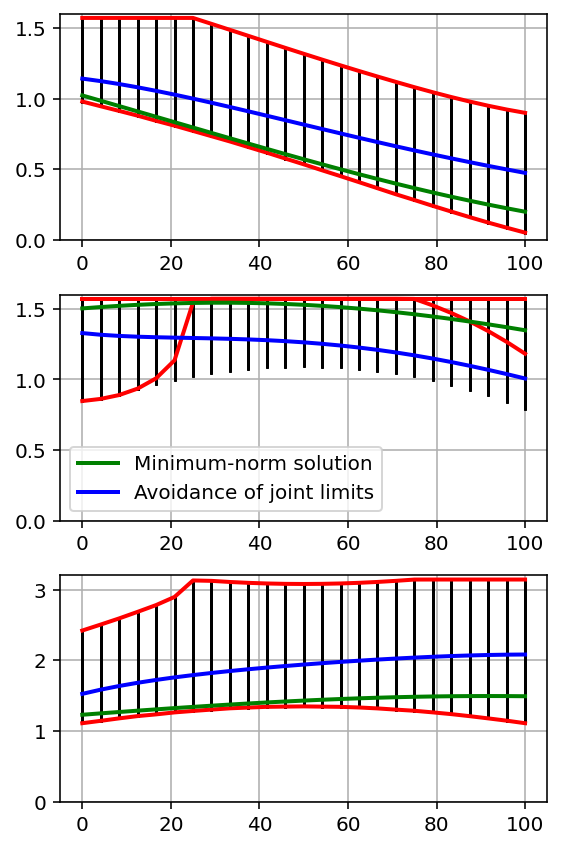}
  \begin{picture}(0,0)
        \put(-180,283){(a) Roadmap of $\theta_1$}
        \put(-180,141){(b) Roadmap of $\theta_2$}
        \put(-180,    0){(c) Roadmap of $\theta_3$}
        \put(-280,  390){$\theta_1$}   
        \put(-280,  251){$\theta_2$}   
        \put(-280,  110){$\theta_3$}
        \put(-248,  363){$P_1$}
        \put(  -25,  375){$P_2$}                   
        \put(-248,  213){$P_1$}
        \put(  -25,  210){$P_2$}            
        \put(-230, 120){$\theta_1^{max}$}
        \put(-139, 60){$\theta_2^{max}$}
        \put(-139, 120){$\theta_2^{max}$}
        \put( -47, 120){$\theta_3^{max}$} 
         \put(-230, 215){$\theta_1^{max}$}
        \put(-139, 262){$\theta_2^{max}$}
        \put( -47, 250){$\theta_3^{max}$}     
         \put(-230, 405){$\theta_1^{max}$}
        \put(-139, 405){$\theta_2^{max}$}
        \put(-139, 330){$\theta_2^{max}$}
        \put( -47, 380){$\theta_3^{max}$}
        \put(-248,   50){$P_1 = (-1, 2)$}
        \put(-65,   50){$P_2 = (+1, 2)$}
        \put( -5,  290){Steps}
        \put( -5,  150){Steps}
        \put( -5,  10){Steps}
  \end{picture}
\caption{Roadmap of a straight-line task from $P_1 =(-1,~2)$ to $P_2=(+1,~2)$ executed by the end-point of a planar 3R manipulator with link lengths $L_1=2$, $L_2=1.5$ and $L_3=1$. The black vertical lines
represent the set of self-motion manifolds associated with the task. Joint limits are shown in red.
The minimum-norm solution is depicted in green, while the joint-limits avoidance strategy
is shown in blue.}  
\label{fig4}   
\end{figure}

Figure \ref{fig3} shows the displacement of the deburring tool around the workpiece when using the proposed  framework. Below, the six-joint positions are shown, between their limits, for both the twist decomposition of eq.(\ref{liguo}) (TWA: blue) and the resolved-motion rate of eq.(\ref{inverse}) (RMR: orange). Apparently, the axes $x$ (red) and $y$ (green) of the tool are automatically rotated around $z$ in order to satisfy the secondary task. For the RMR algorithm, the orientation $x$ or $y$ is set by experience as here, keeping the $y$ coplanar to the $xz$-plane of the robot base. In this situation, the RMR algorithm fails in producing a trajectory within the limits, at index 260 for joint 4 and at index 340 for joint 6. The two reconfigurations occurring at indexes 145 and 315 allow us to avoid the joint limits. 

\section{Conclusions}
\label{sec:5}
A twist decomposition framework for serial manipulators operating under lower-mobility tasks was presented.
By decomposing the twist of the end-effector into redundant and task components, the proposed approach
provides a unified framework for handling both manipulator and task redundancy without explicitly
computing null-space projectors. The resulting inverse kinematics scheme naturally separates
the execution of the primary task from the secondary objectives. The framework effectively handles
both constant and time-varying redundant subspaces. The 3R planar example exhibited a fixed redundancy axis, while the industrial deburring task involved a configuration-dependent redundant subspace. This confirms the ability of the proposed framework to manage diverse redundancy structures consistently.

%
%
%

\end{document}